\documentclass{isprs}
\usepackage{times}
\usepackage{epsfig}
\usepackage{graphicx}
\usepackage[cmex10]{amsmath} 
\usepackage{amssymb}
\usepackage{empheq}
\usepackage{subfig}
\usepackage{pgfplots}
\usepackage{url}
\usepackage{setspace}
\usepackage{geometry} 
\usepackage[labelsep=period]{caption}

\begin{document}

\title{MLPnP - A Real-Time Maximum Likelihood Solution to the Perspective-n-Point Problem}

\author{S. Urban, J.Leitloff, S.Hinz}

\address
{
	Institute of Photogrammetry and Remote Sensing, Karlsruhe Institute of Technology  Karlsruhe \\
	Englerstr. 7, 76131 Karlsruhe, Germany - (steffen.urban, jens.leitloff, stefan.hinz)@kit.edu \\ http://www.ipf.kit.edu
}

\commission{III, }{III}
\workinggroup{III/1} 

\abstract
{
In this paper, a statistically optimal solution to the Perspective-n-Point (PnP) problem is presented.
Many solutions to the PnP problem are geometrically optimal, but do not consider the uncertainties of the observations.
In addition, it would be desirable to have an internal estimation of the accuracy of the estimated rotation and translation parameters of the camera pose.
Thus, we propose a novel maximum likelihood solution to the PnP problem, that incorporates image observation uncertainties and remains real-time capable at the same time.
Further, the presented method is general, as is works with 3D direction vectors instead of 2D image points and is thus able to cope with arbitrary central camera models. 
This is achieved by projecting (and thus reducing) the covariance matrices of the observations to the corresponding vector tangent space.

}


\maketitle

\section{Introduction}\label{sec:Introduction}
The goal of the PnP problem is to determine the absolute pose (rotation and translation) of a calibrated camera in a world reference frame, given known 3D points and corresponding 2D image observations.
The research on PnP has a long history in both the computer vision and the photogrammetry community (here called camera or space resectioning).
Hence, we would first like to emphasize the differences between the definitions used in both communities. 
\paragraph{PnP} Classically in literature, basically two definitions of the problem exist \cite{hu2002note}.
In the first, the \textit{distance based} definition, the problem is formulated in terms of the distances from the projective center to each 3D points, e.g. leading to minimal P3P solutions \cite{haralick1991analysis}.
The second definition is \textit{transformation based}.
Here, the task is to determine the 3D rigid body transformation between object-centered and camera centered coordinate systems, e.g. \cite{fiore2001efficient,horaud1989analytic}.
In both PnP definitions, however, the camera is always assumed to be calibrated and known, allowing to transform image measurements into unit vectors, that point from the camera to the 3D scene points.
More recent approaches extend this classical definition by including unknown camera parameters into the formulation such as the focal length \cite{wu2015p3} or radial distortion \cite{kukelova2013real}.
\paragraph{Camera resectioning} The task in the photogrammetric definition of the problem is to find the projection matrix $\lambda\mathbf{u}=\mathbf{P}_{3\times4}\mathbf{X}$ that transforms homogeneous 3D scene points $\mathbf{X}$ to homogeneous 2D image points $\mathbf{u}$ \cite{hartley2003multiple,luhmann2006close}. The projection matrix is given by $\mathbf{P}_{3\times4} = \mathbf{K}[\mathbf{R}|\mathbf{t}]$ and hence contains the camera matrix $\mathbf{K}$ as well as the rotation $\mathbf{R}$ and translation $\mathbf{t}$.

Thus, fundamentally, the main difference between the definitions in both communities is, that PnP solves only for the absolute pose of the same, calibrated camera.
In camera resectioning the camera is assumed to be unknown and thus part of the formulation of the problem. 
In this paper, we assume the camera to be calibrated and known, thus we present a solution to the Perspective-n-Point problem.

Still, the need for efficient and accurate solutions is driven by a large number of applications.
They range from localization of robots and object manipulation \cite{choi2012robust,collet2009object}, to augmented reality \cite{muller2013mobile} implementations running on mobile devices and having only limited resources, thus focusing on fast solutions.
Especially in industry, surveying or medical environments, involving machine vision, (close-range) photogrammetry \cite{luhmann2006close}, point-cloud registration \cite{weinmann2011fast} and surgical navigation \cite{yang2014vision}, methods are demanded, that are not only robust, but also return a measure of reliability.\par
Even though the research on the PnP problem has a long history, few work has been published on efficient real-time solutions, that take the observation uncertainty into account. 
Most algorithms focus on geometric but not statistic optimality.
To the best of our knowledge, the only work, that includes observation uncertainty into their framework is the Covariant EPPnP of Ferraz et al. \cite{ferraz2014leveraging}.\par
In this paper, we propose a novel formulation of a Maximum Likelihood (ML) solution to the PnP problem.
Further, a general method for variance propagation from image observations to bearing vectors is exploited to avoid singular covariance matrices.
In addition, we benchmark our real-time method against the state-of-the-art and show on a ground truth tracking data set, how our statistical framework can be used, to get a measure of the accuracy of the unknowns given the knowledge about the uncertain observations.
\section{Related Work}\label{sec:RelatedWork}
The minimal number of points to solve the PnP problem is three.
Closed-form solutions to that minimal configuration return up to four solutions, and a fourth point can be use for disambiguation.
Prominent solutions to the P3P problem requiring exactly three points are \cite{kneip2011novel},\cite{li2011stable} and \cite{dementhon1992exact}.
The stability of such algorithms under noisy measurements is limited, hence they are predominately employed in RANSAC \cite{fischler1981random} based outlier rejection schemes.
Apart from the P3P methods, the P4P \cite{fischler1981random,triggs1999camera} and P5P \cite{triggs1999camera}  algorithms exist, still dependent on a fixed number of points.\par
Most solvers, however, can cope with an arbitrary number of feature correspondences.
Basically, they can be categorized into iterative, non-iterative or polynomial, non-polynomial solvers.
Table \ref{tab:methods} lists all methods that are in addition evaluated in the experimental section.\par
Iterative solutions use different objective functions, that are minimized.
In case of LHM \cite{lu2000fast} the pose is initialized using a weak perspective assumption. 
Then they proceed by iteratively minimizing the object space error, i.e. orthogonal deviations between observed ray directions and the corresponding object points in the camera frame.
In the Procrustes PnP \cite{garro2012solving}, the error between the object and the back-projected image points is minimized.
The back-projection is based on the iteratively estimated transformation parameters. 
As iterative methods usually are only guaranteed to find local minima, \cite{schweighofer2008globally} reformulated the PnP problem into a semidefinite programme (SDP).
Despite its O(n) complexity, the runtime of the global optimization method remains tremendous. \par
Likewise, early non-iterative solvers were computational demanding, especially for large point sets.
Among them \cite{ansar2003linear} with O($n^8$), \cite{quan1999linear} with O($n^5$) and \cite{fiore2001efficient} with O($n^2$).
The first efficient non-iterative O($n$) solution is the EPnP by \cite{moreno2007accurate}, that was subsequently extended by \cite{lepetit2009epnp}, employing a fast iterative method to improve the accuracy.
The efficiency comes from the reduction of the PnP problem to finding the position of four control points that are a weighted sum of all 3D points.
After obtaining a linear solution, the weights of the four control points is refined using Gauss-Newton optimization. \par
The most recent, non-iterative state-of-the-art solutions are all polynomial solvers:
The Robust PnP (RPnP) \cite{li2012robust} first splits the PnP problem into multiple P3P problems, that result in a set of fourth order polynomials.
Then the squared sum of those polynomials as well as its derivative is calculated
and the four stationary points are obtained.
The final solution is selected as the stationary point with the smallest reprojection error.
In the Direct-Least-Squares (DLS) \cite{hesch2011direct} method, a nonlinear object space cost function is formulated and polynomial resultant techniques are used to recover the (up to 27) stationary points of a polynomial equation system of fourth order polynomials.
A drawback of this method is the parametrization of the rotation in the cost function by means of the Cayley parameters.
To overcome this, the Accurate and Scalable PnP (ASPnP) \cite{zheng2013aspnp} and the Optimal PnP (OPnP) \cite{zheng2013revisiting} use quaternion based representation of the rotation matrix.
Subsequently, the (up to 40) solutions are found using the Gr\"obner basis technique on the optimality conditions of the algebraic cost function.\par
The linear, non-iterative Unified PnP (UPnP) \cite{kneip2014upnp} goes one step further and integrates the solution to the NPnP (Non-Perspective-N-Point) problem.
The DLS formulation is extended to include non-central camera rays and the stationary points of the first order optimality conditions on the sum of object space errors are found using the Gr\"obner basis methodology.\par
Thus far, all methods assume, that the observations are equally accurate and free of erroneous correspondences.
The first PnP method, that includes an algebraic outlier rejection scheme within the pose estimation, is an extension of the EPnP algorithm called Robust Efficient Procrustes PnP (REPPnP) \cite{ferraz2014very}.
Outliers are removed from the data by sequentially eliminating correspondences that exceed a threshold on an algebraic error.
The procedure remains efficient, as the algorithm operates on the linear system, spanned by the virtual control points of EPnP and thus avoids recalculating the full projection equation in each iteration.
After removing the outliers, the final solution is attained by iteratively solving the closed-form Orthogonal Procrustes problem. 
\par
Yet another extension to the aforementioned EPPnP is termed Covariant EPPnP (CEPPnP) \cite{ferraz2014leveraging}.
It is the first algorithm to inherently incorporate observation uncertainty into the framework.
Again the linear control point system of EPnP is formulated.
Then the covariance information of the feature points is transformed to that space using its Jacobian.
Finally, the Maximum Likelihood (ML) minimization is approximated by an unconstrained Sampson error.\par
In this paper, we propose a new real-time capable, statistically optimal solution to the PnP problem leveraging observation uncertainties.
We propagate 2D image uncertainties to general bearing vectors and show, how a linear ML solution with non-singular covariance matrices can be obtained by using the reduced observation space presented by F\"orstner \cite{forstner2010minimal}.
In contrast to the CEEPnP, where the ML estimator is used, to obtain an estimation of the control point subspace, we optimize directly over the unknown quantities, i.e. rotation and translation and thus, directly obtain pose uncertainties.
Finally, we compare the results of our algorithm to a ground truth trajectory and show, that the estimated pose uncertainties are very close to the ground-truth.
\begin{table}[h]
	\scriptsize
	\begin{center}
		\begin{tabular}{|l|c|c|c|c|}
			\hline
			& iter. & polyn. & using $\mathbf{\Sigma}$ & reference \\ \hline
			LHM & X &  & & \cite{lu2000fast} \\ \hline
			EPnP+GN & (X) &  & & \cite{lepetit2009epnp} \\ \hline
			DLS & & X & & \cite{hesch2011direct} \\ \hline
			RPnP & & X & & \cite{li2012robust} \\ \hline
			PPnP & X &  & & \cite{garro2012solving} \\ \hline
			OPnP & & X & & \cite{zheng2013revisiting} \\ \hline
			ASPnP & & X & & \cite{zheng2013aspnp} \\ \hline
			(R)EPPnP & X &  & & \cite{ferraz2014very} \\ \hline
			CEPPnP & (X) &  & X & \cite{ferraz2014leveraging} \\ \hline
			UPnP & & X & & \cite{kneip2014upnp} \\ \hline
			MLPnP & (X) &  & X & this \\ \hline
		\end{tabular}
	\end{center}
	\caption{Comparison of all tested methods. The methods are categorized into being iterative or a polynomial solver and if they incorporate measurement uncertainty $\mathbf{\Sigma}$. (X) depicts methods, that refine a preliminary result iteratively. }
	\label{tab:methods}
\end{table}
\section{MLPnP}\label{sec:UrbanPnP}
\begin{figure}[h]
	\centering
	\includegraphics[trim = 1mm 70mm 170mm 1mm, clip=true, width=0.49\textwidth]{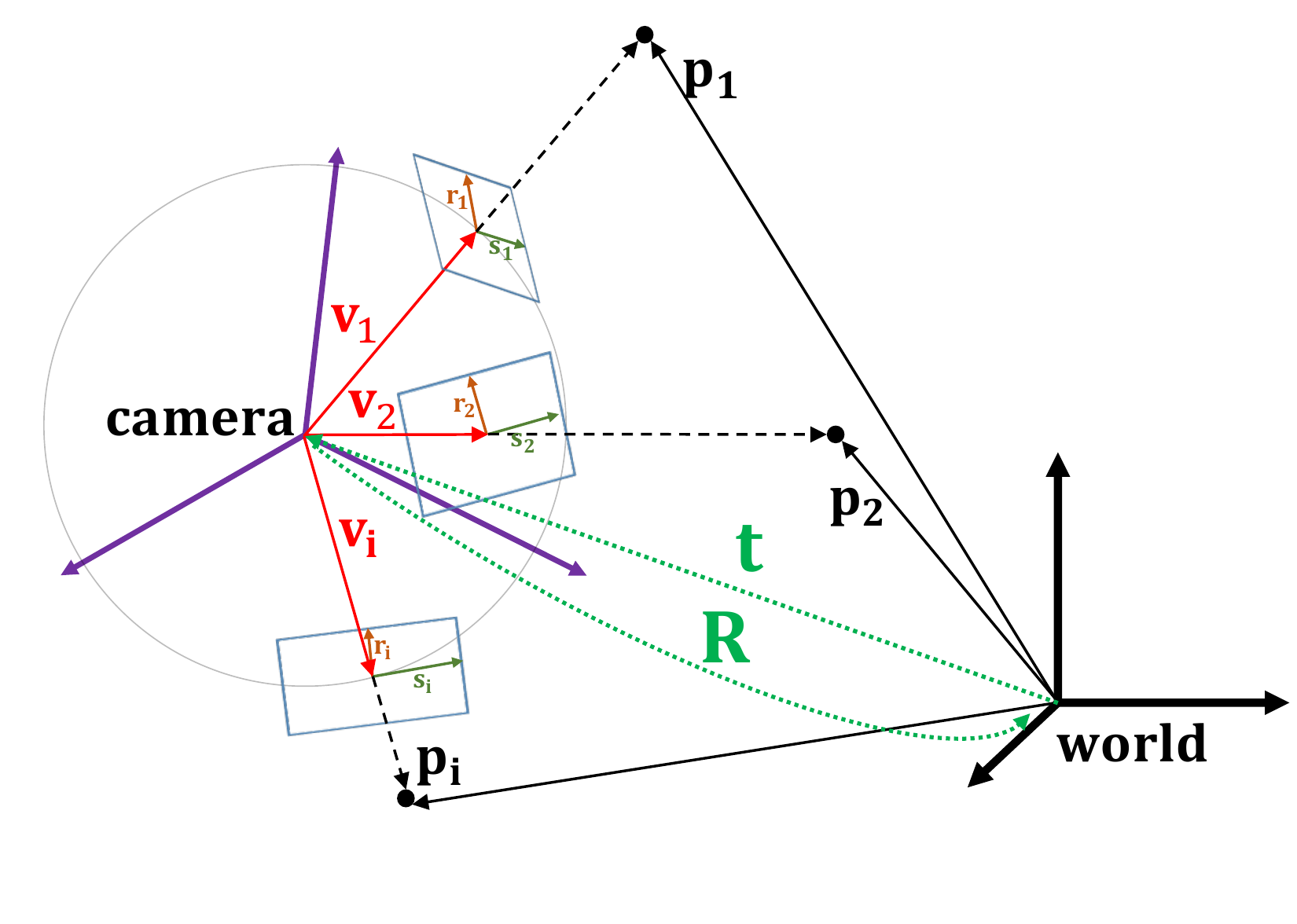}
	\caption{Observations of object points from a camera. The planes at each bearing vector $\mathbf{v}_i$ are spanned by its respective null space vectors $\mathbf{r}_i$ and $\mathbf{s}_i$.}
	\label{fig:UrbanPNPScheme}
\end{figure}

The task of Perspective-n-Point (PnP) is to find the orientation $\mathbf{R} \in SO(3)$ and translation $\mathbf{t} \in \mathbb{R}^3$ that maps the $I$ world points $\mathbf{p}_i, i=1,..,I$ to their corresponding observations $\mathbf{v}_i$ in the camera frame.
This relation is given by:
\begin{equation}
\lambda_i\mathbf{v}_{i} = \mathbf{R}\mathbf{p}_i + \mathbf{t}
\label{eq:pnp}
\end{equation}
where $\lambda_i$ are the depths of each point and the observations $\mathbf{v}_i$ are the measured and thus uncertain quantity having unit length, i.e. $\|\mathbf{v}_i \|=1$.
In the following the methodology of our algorithm is explained.
The parameters and observations are depicted in Fig. \ref{fig:UrbanPNPScheme}.
\subsection{Observations and uncertainty propagation}\label{sec:uncertainty}
Let a world point $\mathbf{p}_i \in \mathbb{R}^3$ be observed by a calibrated camera and let the uncertain observation in the image plane be:
\begin{equation}
\mathbf{x}' =
\begin{bmatrix}	
x' \\ y' 
\end{bmatrix}
,
\mathbf{\Sigma}_{\mathbf{x}'\mathbf{x}'} =
\begin{bmatrix}	
\sigma^2_{x'} & \sigma_{x'y'}\\
\sigma_{y'x'} & \sigma^2_{y'}\\
\end{bmatrix}
\label{eq:imagePlane}
\end{equation}
where the uncertainty about the observed point is described by the 2D covariance matrix $\mathbf{\Sigma}_{\mathbf{x}'\mathbf{x}'}$.
Assuming an arbitrary interior orientation parametrization, the image point $\mathbf{x}'$ is projected to its corresponding three dimensional direction in the camera frame, using the forward projection function $\pi$ (e.g. $\mathbf{K}^{-1}$ in the perspective case):
\begin{equation}
\mathbf{x} =
\pi \mathbf{x}' =
\begin{bmatrix}	
x\\y\\1
\end{bmatrix}
,
\mathbf{J}_{\pi} =
\begin{bmatrix}	
\frac{\partial \pi_{x'}}{\partial x'} & \frac{\partial \pi_{x'}}{\partial y'}\\
\frac{\partial \pi_{y'}}{\partial x'} & \frac{\partial \pi_{y'}}{\partial y'}\\
0 & 0 \\
\end{bmatrix}
\end{equation}
with $\mathbf{J}_{\pi}$ being the Jacobian of the forward projection.
Thus, the uncertainty of the image point $\mathbf{x}'$ is propagated using: 
\begin{equation}
\mathbf{\Sigma}_{\mathbf{x}\mathbf{x}} = 
\mathbf{J}_{\pi}\mathbf{\Sigma}_{\mathbf{x}'\mathbf{x}'}\mathbf{J}_{\pi}^{T}
=
\begin{bmatrix}	
\sigma^2_x & \sigma_{xy} & 0 \\
\sigma_{yx} & \sigma^2_y & 0 \\
0 & 0 & 0 \\
\end{bmatrix}
\end{equation}
where the rank of the covariance matrix $\mathbf{\Sigma}_{\mathbf{x}\mathbf{x}}$ is two, i.e. it is singular and not invertible.
Subsequent spherical normalization yields the final and general observation.
We will refer to them as bearing vectors:
\begin{equation}
\mathbf{v} =
\begin{bmatrix}	
v_x\\v_y\\v_z
\end{bmatrix} =
\frac{\mathbf{x}}{\|\mathbf{x}\|}
,
\mathbf{\Sigma}_{\mathbf{v}\mathbf{v}} =
\begin{bmatrix}	
\sigma^2_{v_{x}} & \sigma_{v_{xy}} & \sigma_{v_{xz}} \\ 
\sigma_{v_{yx}} & \sigma^2_{v_{y}} & \sigma_{v_{yz}} \\
\sigma_{v_{zx}} & \sigma_{v_{zy}} & \sigma^2_{v_{z}} \\
\end{bmatrix}
\label{eq:bearinVectors}
\end{equation}
following \cite{forstner2010minimal} the covariance is propagated using:
\begin{equation}
\mathbf{\Sigma}_{\mathbf{v}\mathbf{v}} = \mathbf{J}\mathbf{\Sigma}_{\mathbf{x}\mathbf{x}}\mathbf{J}^{T}
,
\mathbf{J} = \frac{1}{\|\mathbf{x}\|}(\mathbf{I}_{3} - \mathbf{v}\mathbf{v}^{T})
\label{eq:bearingVecProgragation}
\end{equation}
Observe, that the covariance matrix $\mathbf{\Sigma}_{\mathbf{v}\mathbf{v}}$ remains singular, i.e. a Maximum Likelihood (ML) estimation based on the three residual components of bearing vectors is invalid.
Thus, a minimal representation of the covariance information for the redundant representation of the homogeneous vector $\mathbf{v}$ is desirable.
In the following section, we will introduce the nullspace of vectors and show how this can be subsequently used to get an initial estimate of the absolute orientation of the camera.
\subsection{Nullspace of bearing vectors}\label{sec:nullspace}
The following was developed by \cite{forstner2010minimal}. 
The nullspace of $\mathbf{v}$, spans a two dimensional coordinate system whose axis, denoted as $\mathbf{r}$ and $\mathbf{s}$, are perpendicular to $\mathbf{v}$ and lie in its tangent space:
\begin{equation}
\mathbf{J}_{\mathbf{v}_{r}}(\mathbf{v}) =
null(\mathbf{v}^T) = 
\begin{bmatrix}	
\mathbf{r} & \mathbf{s}
\end{bmatrix}
=
\begin{bmatrix}	
r_1 & s_1 \\
r_2 & s_2 \\
r_3 & s_3 \\
\end{bmatrix}
\label{eq:nullspaceVecs}
\end{equation}
The function $null(\cdot)$ calculates the Singular Value Decomposition (SVD) of $\mathbf{v}$ and takes the two eigenvectors corresponding to the two zero eigenvalues.
Further we assume $\mathbf{J}_{\mathbf{v}_{r}}$ to be an orthonormal matrix, i.e. $\mathbf{J}^T_{\mathbf{v}_{r}}(\mathbf{v})\mathbf{J}_{\mathbf{v}_{r}}(\mathbf{v}) = \mathbf{I}_2$.
Note, that $\mathbf{J}_{\mathbf{v}_{r}}$ in addition represents the Jacobian of the transformation from the tangent space to the original vector.
Thus the transpose $\mathbf{J}^T_{\mathbf{v}_{r}}$ yields the transformation from the original homogeneous vector $\mathbf{v}$ to its reduced equivalent $\mathbf{v}_r$.
\begin{equation}
\mathbf{v}_r = 
\begin{bmatrix}
dr \\ ds
\end{bmatrix} =
\mathbf{J}^T_{\mathbf{v}_{r}}(\mathbf{v})\mathbf{v} = \mathbf{0}
\label{eq:projectVtoVr}
\end{equation}
with nonsingular covariance
\begin{equation}
\mathbf{\Sigma}_{\mathbf{v}_r\mathbf{v}_r} 
=
\mathbf{J}^T_{\mathbf{v}_{r}}(\mathbf{v})\mathbf{\Sigma}_{\mathbf{v}\mathbf{v}}\mathbf{J}_{\mathbf{v}_{r}}(\mathbf{v})
=
\begin{bmatrix}	
\sigma^2_{v_{r_{x}}} & \sigma_{v_{r_{xy}}}\\ 
\sigma_{v_{r_{xy}}} & \sigma^2_{v_{r_{y}}}
\end{bmatrix}
\end{equation}
Another way to think about $\mathbf{v}_r$ is as a residual in the tangent space.
In the following, we will exploit Eq. \ref{eq:projectVtoVr} to get a linear estimate of the rotation and translation of the camera in the world frame by minimizing this residual in the tangent space.
\subsection{Linear estimation of the camera pose}\label{sec:linearcampose}
Using Eq.\ref{eq:pnp} and \ref{eq:nullspaceVecs} we can reformulate Eq. \ref{eq:projectVtoVr}:
\begin{equation}
\begin{bmatrix}
dr \\ ds
\end{bmatrix} =
\begin{bmatrix}
\mathbf{r}^T \\
\mathbf{s}^T
\end{bmatrix}
\lambda^{-1}_{i}(\mathbf{R}\mathbf{p}_i + \mathbf{t})
=
\mathbf{0}
\label{eq:mustbezero}
\end{equation}
with $\lambda_{i} \neq 0$.
Thus, if we knew the absolute orientation of our camera, the projection of a world point $p_i$ to the tangent space of $\mathbf{v}$, should result in the same reduced coordinates, i.e. zero residual.
Expanding Eq. \ref{eq:mustbezero} yields:
\begin{equation}
\begin{split}
0 = r_1(\hat{r}_{11}p_x+\hat{r}_{12}p_y+\hat{r}_{13}p_z+\hat{t}_1) \\
+r_2(\hat{r}_{21}p_x+\hat{r}_{22}p_y+\hat{r}_{23}p_z+\hat{t}_2) \\
+r_3(\hat{r}_{31}p_x+\hat{r}_{32}p_y+\hat{r}_{33}p_z+\hat{t}_3)
\\
0 = s_1(\hat{r}_{11}p_x+\hat{r}_{12}p_y+\hat{r}_{13}p_z+\hat{t}_1) \\
+s_2(\hat{r}_{21}p_x+\hat{r}_{22}p_y+\hat{r}_{23}p_z+\hat{t}_2) \\
+s_3(\hat{r}_{31}p_x+\hat{r}_{32}p_y+\hat{r}_{33}p_z+\hat{t}_3)
\end{split}                               
\end{equation}
with $\mathbf{p} = [p_x,p_y,p_z]^T$.
Now, both equations are linear in the unknowns, i.e. we can stack them in a design matrix $\mathbf{A}$ to obtain a homogeneous system of linear equations:
\begin{equation}
\mathbf{A}\mathbf{u} = \mathbf{0}
\label{eq:homoEQ}
\end{equation}
with $\mathbf{u} = [\hat{r}_{11},\hat{r}_{12},\hat{r}_{13},\hat{r}_{21},\hat{r}_{22},\hat{r}_{23},\hat{r}_{31},\hat{r}_{32},\hat{r}_{33},\hat{t}_1,\hat{t}_2,\hat{t}_3]^T$.
As each observation yields two residuals, at least $I>5$ points are necessary to solve Eq. \ref{eq:homoEQ}.
Assuming uncorrelated observations, the stochastic model is given by:
\begin{equation}
\mathbf{P} = 
\begin{bmatrix}
\Sigma^{-1}_{\mathbf{v}^1_r\mathbf{v}^1_r} & \hdots & 0 \\
 \vdots & \ddots &  \vdots \\
 0 & \hdots & \Sigma^{-1}_{\mathbf{v}^i_r\mathbf{v}^i_r} 
\end{bmatrix}
\end{equation}
and the final normal equations are:
\begin{equation}
\mathbf{A}^T\mathbf{P}\mathbf{A}\mathbf{u} = \mathbf{N}\mathbf{u} = \mathbf{0}
\label{eq:normaleq}
\end{equation}
We find the $\mathbf{u}$ that minimizes Eq. \ref{eq:normaleq} subject to $\|\mathbf{u}\| = 1$,
using SVD:
\begin{equation}
\mathbf{N} = \mathbf{U}\mathbf{D}\mathbf{V}^T
\label{eq:svd}
\end{equation}
The solution is the particular column of $\mathbf{V}$ that corresponds to the smallest singular value in $\mathbf{D}$:
\begin{equation}
\hat{\mathbf{R}} = 
\begin{bmatrix}
\hat{r}_{11} & \hat{r}_{12} & \hat{r}_{13} \\ \hat{r}_{21} & \hat{r}_{22} & \hat{r}_{23} \\ \hat{r}_{31} & r_{32} & r_{33}
\end{bmatrix} ,
\mathbf{t} = 
\begin{bmatrix}
\hat{t}_1,\hat{t}_2,\hat{t}_3
\end{bmatrix}
\end{equation}
It is determined up to a scale factor, thus the translational part $\hat{\mathbf{t}}$ only points in the right direction.
The scale can be recovered from the fact, that the norm of each column $\hat{\mathbf{r}}_1$, $\hat{\mathbf{r}}_2$ and $\hat{\mathbf{r}}_3$ of the rotation matrix $\hat{\mathbf{R}}$ must equal one.
Hence, the final translation is:
\begin{equation}
\mathbf{t} = \frac{\hat{\mathbf{t}}}{\sqrt[3]{\|\hat{\mathbf{r}}_1\| \|\hat{\mathbf{r}}_2\| \|\hat{\mathbf{r}}_3\|}}
\end{equation}
The exploited constrain shows, that the 9 rotation parameters do not define a correct rotation matrix.
This, can be solved by calculating the SVD of $\hat{\mathbf{R}}$:
\begin{equation}
\hat{\mathbf{R}}
=
\mathbf{U}_R\mathbf{D}_R\mathbf{V}^T_R
\end{equation}
and the best rotation matrix minimizing the Frobenius norm is found as:
\begin{equation}
\mathbf{R} = \mathbf{U}_R\mathbf{V}^T_R.
\label{eq:rotationFrobenius}
\end{equation}
Up to this point, a linear ML estimation of the absolute camera pose is obtained.
To increase the accuracy, a non-linear refinement procedure is used.
Doing a subsequent refinement of the initial estimate is a common procedure, e.g. performed in \cite{lepetit2009epnp}, \cite{ferraz2014very} or \cite{lu2000fast}.
\subsection{Non-linear refinement}\label{sec:nonlinearcampose}
We apply a Gauss-Newton optimization to iteratively refine the camera pose.
Specifically, we minimize the tangent space residuals defined in Eq. \ref{eq:mustbezero}.
This is reasonably fast for two reasons.
On the one hand, the nullspace vectors are already calculated and we simply have to calculate the dot products between the tangent space vectors and each transformed world point.
On the other hand, the results of the linear estimates are already close to a local minimum, i.e. the Gauss-Newton optimization converges quickly.
In practice we found, that a maximum number of five iterations is sufficient.
To arrive at a minimal representation of the rotation matrix, we chose to express $\mathbf{R}$ in terms of the Rodriguez parametrization.
\subsection{Planar case}\label{sec:planarcase}
In the planar case, the SVD (Eq.\ref{eq:svd}) yields up to four solution vectors, as the corresponding singular values become small (close to zero).
In this case, the solution is a linear combination of those vectors.
To solve for the coefficients, an equation system with non-linear constraints had to be solved.
We avoid this using the following trick.
Let $\mathbf{M} = [\mathbf{p}_1,\mathbf{p}_2,..,\mathbf{p}_i]$ be a 3 $\times$ $I$ matrix of all world points.
The eigenvalues of $\mathbf{S}=\mathbf{M}\mathbf{M}^T$ give us information about the distribution of the world points.
In the ordinary 3D case, the rank of matrix $\mathbf{S}$ is three and the smallest eigenvalue is not close to zero.
In the planar case, the smallest eigenvalue becomes small and the rank of matrix $\mathbf{S}$ is two.
If the world points lie on a plane that is spanned by two coordinate axis, respectively, i.e. one of the elements of all world points is a constant, we could simply omit the corresponding column from matrix $\mathbf{A}$ and get a distinct solution.\par
In general, the points can lie on an arbitrary plane in the world frame.
Thus, we use the eigenvectors of $\mathbf{S}$ as a rotation matrix $\mathbf{R}_S$ and rotate the world points to a new frame using:
\begin{equation}
\hat{\mathbf{p}}_i = \mathbf{R}^T_S\mathbf{p}_i
\end{equation}
Here, we can identify the constant element of the coordinates and omit the corresponding column from the design matrix $\mathbf{A}$.
Note, that this transformation does not change the structure of the problem.
The rotation matrix obtained after SVD (Eq. \ref{eq:rotationFrobenius}), simply has to be rotated back to the original coordinate frame:
\begin{equation}
\mathbf{R} = \mathbf{R}_S\mathbf{R}
\end{equation}
To keep our method as general as possible, the matrix $\mathbf{S}$ is always calculated.
We then simply switch between the planar and the ordinary case, depending on the rank of the matrix $\mathbf{S}$.
To determine the rank we use rank-revealing QR decomposition with full pivoting and set a threshold (1e-10) on the smallest eigenvalue.

\section{Results}\label{sec:Results}
In this section, we compare our algorithm to all state-of-the-art algorithms using synthetic as well as real data.
To reasonably assess the runtime performance of each algorithm, we categorize the state-of-the-art solvers according to their implementation:
\begin{description}
	\item[Matlab]\hfill \\
	OPnP \cite{zheng2013revisiting}, EPPnP \cite{ferraz2014very}, CEPPnP \cite{ferraz2014leveraging}, DLS \cite{hesch2011direct}, \\LHM \cite{lu2000fast},
	ASPnP \cite{zheng2013aspnp}, SDP \cite{schweighofer2008globally}, RPnP \cite{li2012robust},\\PPnP \cite{garro2012solving}
	\item[C++,mex]\hfill \\ 
	UPnP \cite{kneip2014upnp}, EPnP+GN \cite{lepetit2009epnp}, MLPnP (this paper)
\end{description}
Note however, that the Matlab implementations of many algorithms are already quite optimized, and it is unclear how large the performance increase of a C++ version would be. \par           
Our method is implemented in both Matlab and C++.
We integrated the C++ version in OpenGV \cite{kneip2014opengv}.
The Matlab version will be made publicly available from the website of the corresponding author, i.e. all results are reproducible.
All experiments have been conducted on a Laptop with Intel Core i7-3630QM@2.4Ghz.
\subsection{Synthetic experiments}\label{sec:syntheticExperiments}
We use the Matlab toolbox provided by the authors of \cite{ferraz2014leveraging,li2012robust,zheng2013revisiting} and compare our algorithm in terms of accuracy and speed. The simulation configurations as well as the evaluation metrics are also part of the toolbox and are briefly described in the following. All experiments are repeated $T=250$ times and the mean pose errors are reported. \par
Assume a virtual calibrated camera with a focal length of $f=$ 800 pixels.
First, 3D points are randomly sampled in the camera frame on the interval [-2, 2]$\times$[-2, 2]$\times$[4,8] and projected to the image plane. 
Then, the image plane coordinates $\mathbf{x}'$ are perturbed by Gaussian noise with different standard deviations $\sigma$ yielding a covariance matrix $\Sigma_{\mathbf{x}'\mathbf{x}'}$ for each feature.
At this point, we arrived at Eq. \ref{eq:imagePlane}. \par
Now, the image plane coordinates $\mathbf{x}'$ are usually transformed to normalized image plane coordinates applying the forward projection $\pi$.
In the synthetic experiment, this simplifies to perspective division: $\mathbf{x} = [x'/f,y'/f,1]^T$.
Most algorithms work with the first two elements of $\mathbf{x}$, i.e. normalized image plane coordinates.
Instead, we apply Eq. \ref{eq:bearinVectors} and spherically normalize the vectors, to simulate a general camera model.
Further, we propagate the covariance information using Eq. \ref{eq:bearingVecProgragation}. \par
Finally, the ground truth translation vector $\mathbf{t}_{gt}$ is chosen as the centroid of the 3D points and the ground truth rotation $\mathbf{R}_{gt}$ is randomly sampled.
Subsequently, all points are transformed to the world frame yielding $\mathbf{p}_i$.\par
The rotation accuracy in degree between $\mathbf{R}_{gt}$ and $\mathbf{R}$ is measured as $\max^3_{k=1}(\arccos(\mathbf{r}^T_{k,gt}\cdot\mathbf{r}_k)\times180/\pi)$, where $\mathbf{r}_{k,gt}$ and $\mathbf{r}$ are the k-th column of the respective rotation matrix.
The translation accuracy is computed in \% as $\|\mathbf{t}_{gt}-\mathbf{t}\|/\|\mathbf{t}\|\times 100$. \par
The first column of \ref{fig:ErrorOrdinary3D_sigma} depicts the ordinary and the second column the planar case, where the $Z$ coordinate of the world points is set to zero.
The first two rows of Fig. \ref{fig:ErrorOrdinary3D_sigma} depict the mean accuracy for rotation and translation for an increasing number of points (I=10,..,200).
Following the experiments of \cite{ferraz2014leveraging}, the image plane coordinates are perturbed by Gaussians with $\sigma$ = [1,..,10] and 10\% of the features are perturbed by each noise level, respectively. \par
For the third and fourth row of Fig. \ref{fig:ErrorOrdinary3D_sigma} the number of features is kept constant (I=50) and only the noise level is increased.
This time, a random $\sigma$ is chosen for each feature from 0 to the maximum number, ranging from 0 to 10.\par
The experiments for the ordinary case show, that MLPnP outperforms all other state-of-the-art algorithms in terms of accuracy.
Moreover, it stays among the fastest algorithms as depicted in Fig. \ref{fig:time}.
For less than 20 points, the algorithm is even faster than EPnP, that is still the fastest PnP solution.\par
\input{SimulationResults}
\subsection{Real Data}\label{sec:realData}
\begin{figure}
	\centering
	\subfloat[\label{fig:cameraSystem}] 
	{
		\includegraphics[ height=0.3\textwidth]{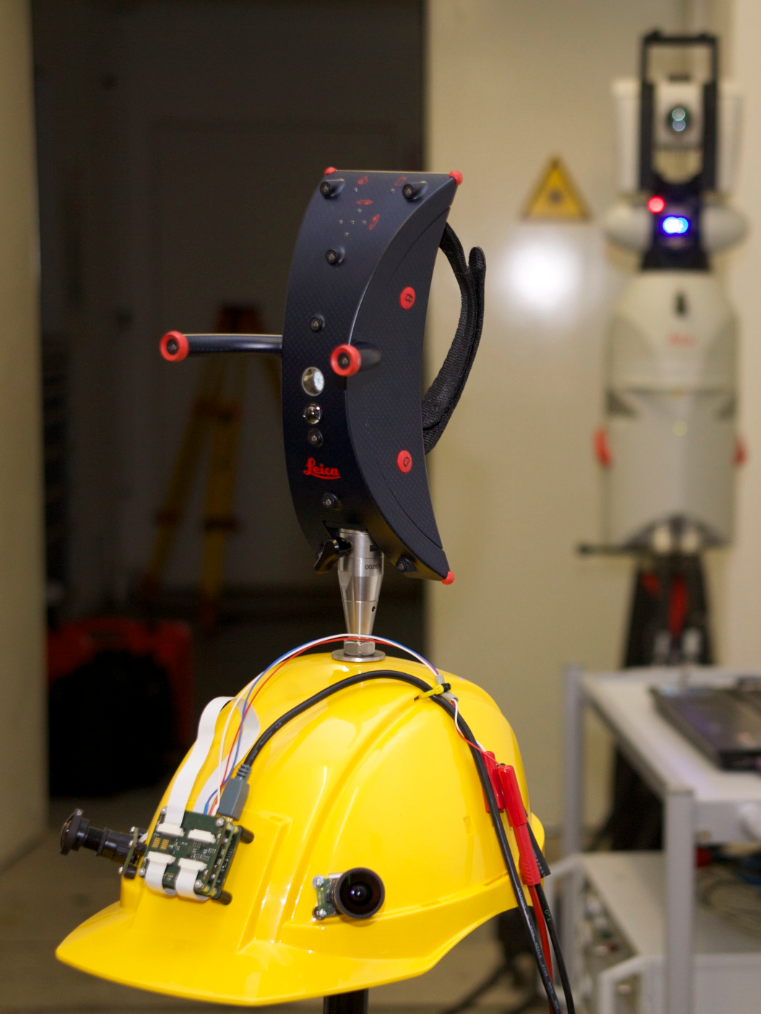}
	}
	\subfloat[\label{fig:trajImages}] 
	{
		\includegraphics[ height=0.3\textwidth]{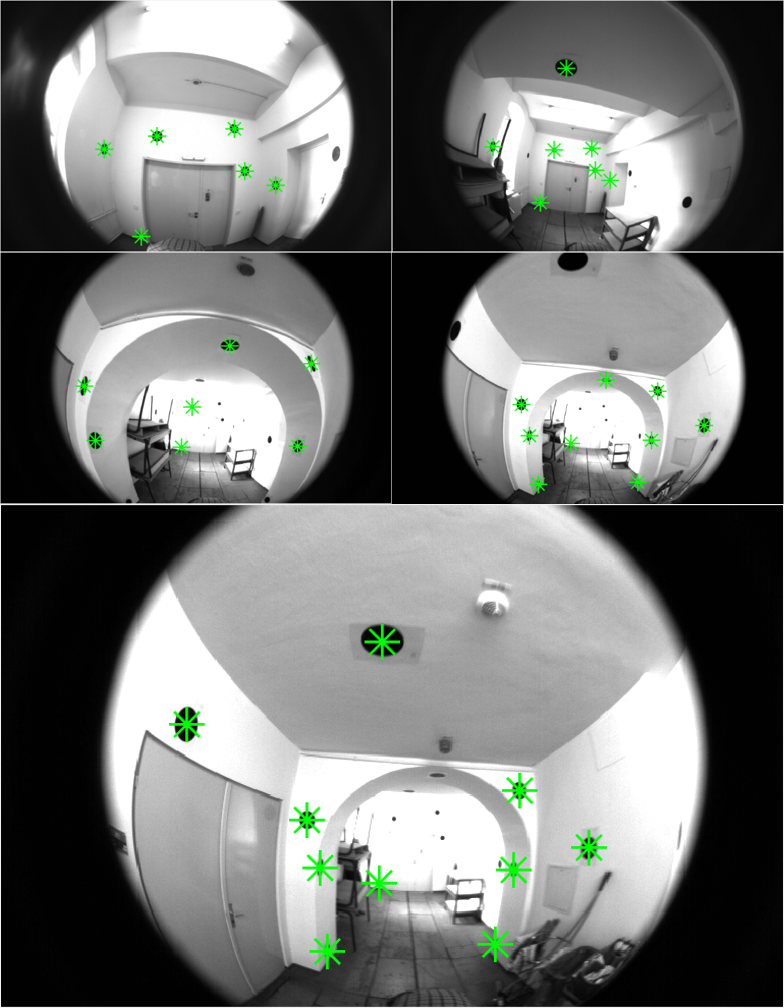}
	}
	\caption{(a) Camera system with mounted T-Probe. The background shows the lasertracker. (b) Some images taken from the trajectory. Tracked world points are depicted in green.}
\end{figure}

To evaluate the performance of our proposed MLPnP method on real data, we recorded a trajectory of T = 200 poses of a moving, head mounted fisheye camera in an indoor environment.
The camera is part of a helmet multi-camera system depicted in Fig. \ref{fig:cameraSystem}.
We use the camera model of \cite{scaramuzza2006flexible} and the toolbox of \cite{urban2015improved} to calibrate the camera.
All sensors have a resolution of 754$\times$480 pixels and are equipped with equal fisheye lenses.
The field of view of each lens covers $185^\circ$.
The ground truth poses $\mathbf{M}^{t}_{gt}$ are obtained by a Leica T-Probe, mounted onto the helmet and tracked by a Leica lasertracker.
This tracker was in addition used to measure the 3D coordinates of the world points.
The position accuracy of this system is $\sigma_{pos} \approx 100 \mu m = 0.1mm$ and the angle accuracy is indicated with $\sigma_{angle} \approx 0.05mrad$.\par
For each frame $t$ in the trajectory, the epicenters of $I$ passpoints (planar, black filled circles) are tracked over time, as depicted in Fig. \ref{fig:trajImages}.
The number of passpoints $I$, that are visible in each frame ranges from 6 to 13.\par
To assess the quality of the estimated camera pose, we calculate the mean standard deviation (SD) $\bar{\sigma}_{\mathbf{R}}$ and  $\bar{\sigma}_{\mathbf{t}}$ of the relative orientation between camera and T-Probe.
The relative orientation for each frame is calculated as follows:
\begin{equation}
\mathbf{M}_{rel} =  \mathbf{M}_{gt}^{-1}\mathbf{M}_t
\end{equation}
where $\mathbf{M}_t = [\mathbf{R},\mathbf{t}]$ is the current camera pose and is estimated by each of the 13 algorithms respectively.\par
\subsection{Including Covariance Information}\label{sec:IncludeCovInfo}
For the first frame $t=1$, the covariance information about the measurements $\mathbf{\Sigma}_{\mathbf{x}'\mathbf{x}'}$ is set to the identity $\mathbf{I}_2$ for each feature.
To this point, the only thing we know about the features is, that they are measured with identical accuracy of 1 pixel.
For the following frame $t=2$, however, estimation theory can tell us, how well a point was measured in the previous frame, given the estimated parameter vector. \par
\begin{figure*}
\subfloat[\label{fig:realDataOri}]
{
	\begin{tikzpicture}[baseline=(current axis.south)]
		\begin{axis}[
			grid=major,
			bar width = 5pt,
			ymax=0.5,ymin=0,
			xmax = 14, xmin = 0,
			ylabel={\scriptsize mean orientation SD $\bar{\sigma}_{\mathbf{R}}$ [deg]},	
			yticklabel pos=left,
 			width = 0.33\textwidth,
 			height = 0.25\textwidth,
 			x tick label style={rotate=-45, anchor=west, align=right}, 
            xtick={1,2,3,4,5,6,7,8,9,10,11,12,13},
            xticklabels =  {\tiny MLPnP,\tiny MLPnP+$\Sigma$,\tiny LHM,\tiny EPnP+GN,\tiny RPnP,\tiny DLS,\tiny PPnP,\tiny ASPnP,\tiny SDP,\tiny OPnP,\tiny EPPnP,\tiny CEPPnP,\tiny UPnP}]
            \addplot[mark=none, black, samples=2] 
            	table [x index={0},y index={1}] {line_ori.txt};
            \addplot[mark=none, black, samples=2] {0.12381};
	    	\addplot[fill=green,ybar]	    
	    	table [x expr=1,y=MLPnP] {std_ori.txt};
    		\addplot[fill=green,ybar]
    	    table [x expr=2,y=MLPnP+cov]{std_ori.txt}; 
    	    \addplot[fill=black,ybar]
    	    table [x expr=3,y=LHM]{std_ori.txt};
    	    \addplot[fill=magenta,ybar]    
    	    table [x expr=4,y=EPnP+GN]{std_ori.txt};
    	    \addplot[fill=teal,ybar]           
    	    table [x expr=5,y=RPnP]{std_ori.txt}; 
    	    \addplot[fill=gray,ybar]    
    	    table [x expr=6,y=DLS] {std_ori.txt};
    	    \addplot[fill=blue,ybar]
    	    table [x expr=7,y=PPnP]{std_ori.txt}; 
    	    \addplot[fill=cyan,ybar]
    	    table [x expr=8,y=ASPnP]{std_ori.txt};	
    	    \addplot[fill=orange,ybar]
    	    table [x expr=9,y=SDP]{std_ori.txt};
    	    \addplot[fill=pink,ybar]                            
    	    table [x expr=10,y=OPnP]{std_ori.txt};
    	    \addplot[fill=purple,ybar]
    	    table [x expr=11,y=EPPnP] {std_ori.txt};
    	    \addplot[fill=red,ybar]
    	    table [x expr=12,y=CEPPnP] {std_ori.txt};
    	    \addplot[fill=violet,ybar]
    	    table [x expr=13,y=UPnP]{std_ori.txt}; 	   
		\end{axis}
	\end{tikzpicture}
}
\subfloat[\label{fig:realDataPos}]
{
	\begin{tikzpicture}[baseline=(current axis.south)]
	\begin{axis}[
	grid = major,
	bar width = 5pt,
	ymax=2,ymin=0,	
	xmax = 14, xmin = 0,
	ylabel={\scriptsize mean translation SD $\bar{\sigma}_{\mathbf{t}}$ [cm]},	
	yticklabel pos=left,
	width = 0.33\textwidth,
	height = 0.25\textwidth,
	x tick label style={rotate=-45, anchor=west, align=left}, 
	xtick={1,2,3,4,5,6,7,8,9,10,11,12,13},
	xticklabels =  {\tiny MLPnP,\tiny MLPnP+$\Sigma$,\tiny LHM,\tiny EPnP+GN,\tiny RPnP,\tiny DLS,\tiny PPnP,\tiny ASPnP,\tiny SDP,\tiny OPnP,\tiny EPPnP,\tiny CEPPnP,\tiny UPnP}]
	\addplot[mark=none, black, samples=2] 
	table [x index={0},y index={1}] {line_pos.txt};
	    	\addplot[fill=green,ybar]	    
	    	table [x expr=1,y=MLPnP] {std_pos.txt};
	    	\addplot[fill=green,ybar]
	    	table [x expr=2,y=MLPnP+cov]{std_pos.txt}; 
	    	\addplot[fill=black,ybar]
	    	table [x expr=3,y=LHM]{std_pos.txt};
	    	\addplot[fill=magenta,ybar]    
	    	table [x expr=4,y=EPnP+GN]{std_pos.txt};
	    	\addplot[fill=teal,ybar]           
	    	table [x expr=5,y=RPnP]{std_pos.txt}; 
	    	\addplot[fill=gray,ybar]    
	    	table [x expr=6,y=DLS] {std_pos.txt};
	    	\addplot[fill=blue,ybar]
	    	table [x expr=7,y=PPnP]{std_pos.txt}; 
	    	\addplot[fill=cyan,ybar]
	    	table [x expr=8,y=ASPnP]{std_pos.txt};	
	    	\addplot[fill=orange,ybar]
	    	table [x expr=9,y=SDP]{std_pos.txt};
	    	\addplot[fill=pink,ybar]                            
	    	table [x expr=10,y=OPnP]{std_pos.txt};
	    	\addplot[fill=purple,ybar]
	    	table [x expr=11,y=EPPnP] {std_pos.txt};
	    	\addplot[fill=red,ybar]
	    	table [x expr=12,y=CEPPnP] {std_pos.txt};
	    	\addplot[fill=violet,ybar]
	    	table [x expr=13,y=UPnP]{std_pos.txt}; 
	\end{axis}
	\end{tikzpicture}
}
\subfloat[\label{fig:realDataTime}]
{
	\begin{tikzpicture}[baseline=(current axis.south)]
	\begin{axis}[
	grid=major,
	bar width = 5pt,
	xmax = 14, xmin = 0,
	ymax=8,ymin=0,
	ylabel={\scriptsize mean runtime [ms]},
	yticklabel pos=left,
	width = 0.33\textwidth,
	height = 0.25\textwidth,
	x tick label style={rotate=-45, anchor=west, align=left}, 
	xtick={1,2,3,4,5,6,7,8,9,10,11,12,13},
	xticklabels =  {\tiny MLPnP,\tiny MLPnP+$\Sigma$,\tiny LHM,\tiny EPnP+GN,\tiny RPnP,\tiny DLS,\tiny PPnP,\tiny ASPnP,\tiny SDP,\tiny OPnP,\tiny EPPnP,\tiny CEPPnP,\tiny UPnP}]
	\addplot[mark=none, black, samples=2] 
	table [x index={0},y index={1}]{line_run.txt};
	    	\addplot[fill=green,ybar]	    
	    	table [x expr=1,y=MLPnP] {mean_run.txt};
	    	\addplot[fill=green,ybar]
	    	table [x expr=2,y=MLPnP+cov]{mean_run.txt}; 
	    	\addplot[fill=black,ybar]
	    	table [x expr=3,y=LHM]{mean_run.txt};
	    	\addplot[fill=magenta,ybar]    
	    	table [x expr=4,y=EPnP+GN]{mean_run.txt};
	    	\addplot[fill=teal,ybar]           
	    	table [x expr=5,y=RPnP]{mean_run.txt}; 
	    	\addplot[fill=gray,ybar]    
	    	table [x expr=6,y=DLS] {mean_run.txt};
	    	\addplot[fill=blue,ybar]
	    	table [x expr=7,y=PPnP]{mean_run.txt}; 
	    	\addplot[fill=cyan,ybar]
	    	table [x expr=8,y=ASPnP]{mean_run.txt};	
	    	\addplot[fill=orange,ybar]
	    	table [x expr=9,y=SDP]{mean_run.txt};
	    	\addplot[fill=pink,ybar]                            
	    	table [x expr=10,y=OPnP]{mean_run.txt};
	    	\addplot[fill=purple,ybar]
	    	table [x expr=11,y=EPPnP] {mean_run.txt};
	    	\addplot[fill=red,ybar]
	    	table [x expr=12,y=CEPPnP] {mean_run.txt};
	    	\addplot[fill=violet,ybar]
	    	table [x expr=13,y=UPnP]{mean_run.txt}; 
	\end{axis}
	\end{tikzpicture}
}
\caption{(a) mean orientation and (b) mean translation standard deviation over all frames of the real trajectory. (c) mean runtime over all frames. CEPPnP and MLPnP+$\mathbf{\Sigma}$ incorporate measurement uncertainty. All other algorithms assume equally well measured image points.}
\label{fig:realData}
\end{figure*}
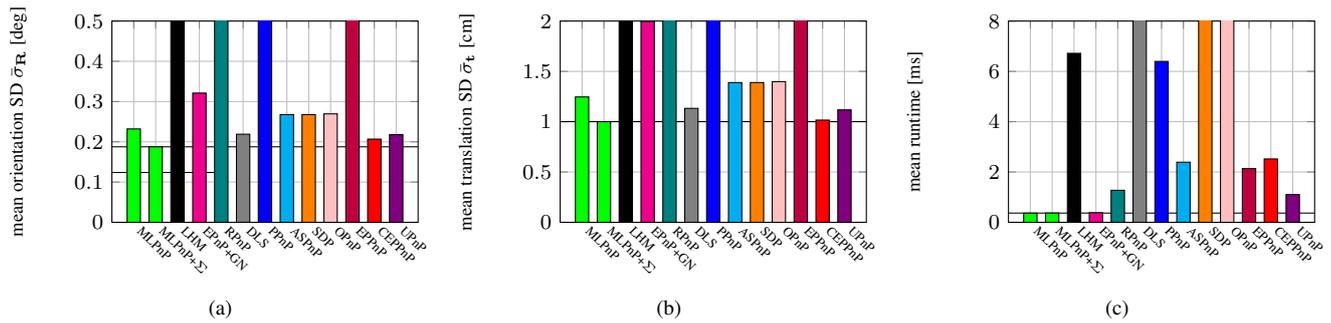

Let $\mathbf{A}$ be the Jacobian of our residual function Eq. \ref{eq:mustbezero}.
Then, the covariance matrix of the unknown rotation and translation parameters is:
\begin{equation}
\mathbf{\Sigma}_{\hat{\mathbf{r}}\hat{\mathbf{t}}} = (\mathbf{A}^T\mathbf{P}\mathbf{A})^{-1}
\label{eq:covMatParams}
\end{equation}
and cofactor matrix of the observations:
\begin{equation}
\mathbf{Q}_{\mathbf{v}_r\mathbf{v}_r} = \mathbf{A}\mathbf{\Sigma}_{\hat{\mathbf{r}}\hat{\mathbf{t}}}\mathbf{A}^T
\end{equation}
Observe, that this gives us the cofactor matrix in the reduces observation space.
Thus, we project them back to the true observation space of the respective point i:
\begin{equation}
\mathbf{\Sigma}_{\mathbf{v}_i\mathbf{v}_i} =
\mathbf{J}_{\mathbf{v}^i_{r}}(\mathbf{v}_i) \mathbf{Q}_{\mathbf{v}^i_r\mathbf{v}^i_r} \mathbf{J}^T_{\mathbf{v}^i_{r}}(\mathbf{v}_i) 
\end{equation}
Now, we can set the covariance matrices $\mathbf{\Sigma}_{\mathbf{v}_i\mathbf{v}_i}$ for frame $t=2$ to the estimated values from the first frame $t=1$ and so on.
The results from this method are depicted as MLPnP+$\Sigma$ in Fig. \ref{fig:realData}.
Clearly, the consistent use of the full covariance information gives another performance increase.%
Fig. \ref{fig:realDataOri} shows the mean standard deviation of the three orientation components.
\subsection{External vs. internal accuracies}\label{sec:GTvsET}
Usually, it is desirable to have a measure of reliability about the result of an algorithm.
A common approach in geometric vision, is to quantify the quality of the pose in terms of the reprojection error.
Most of the time, this might be sufficient.
Many computer vision systems are based on geometric algorithms and
need a measure of how accurate and robust objects are mapped in the camera images.
But sometimes it is useful to get more information about the quality of the estimated camera pose, e.g. for probabilistic SLAM systems.
As MLPnP is a full ML estimator, we can employ estimation theory to obtain the covariance information about the estimated pose parameters and thus their standard deviations as a measure of reliability.
Let $\mathbf{r}$ be the vector of stacked residuals $[dr_i,ds_i]^T$.
First, we calculate the variance factor:
\begin{equation}
\sigma^2_0 = \frac{\mathbf{r}^T\mathbf{P}\mathbf{r}}{b}
\end{equation}
with $\mathbf{P}$ being the stochastic model and $b = 2I-6$ is the redundancy of our problem.
Then, the 6$\times$6 covariance matrix $\mathbf{\Sigma}_{\hat{\mathbf{r}}\hat{\mathbf{t}}}$ of the unknowns is extracted with Eq. \ref{eq:covMatParams}.
Finally, the 6$\times$1 vector of standard deviations of the camera pose is:
\begin{equation}
\mathbf{\sigma}_{\hat{\mathbf{r}},\hat{\mathbf{t}}} = \sigma_0\sqrt{diag(\mathbf{\Sigma}_{\hat{\mathbf{r}}\hat{\mathbf{t}}})}
\end{equation}
We extract the covariance matrix for every frame in the trajectory and calculate the mean standard deviation over all frames.
Tab. \ref{tab:uncertainty} depicts the results compared to the standard deviation that was calculated to the ground truth.
The marginal differences between the internal uncertainty estimations and the external ground truth uncertainties is an empirical proof that our proposed ML estimator is statistically optimal.
\begin{table}[h]
	\begin{center}
	\begin{tabular}{|c|c|c|}
		\hline
		& $\sigma_{angle}$ [deg] & $\sigma_{pos}$ [cm] \\ \hline
		ground truth & 0.19 & 1.03 \\ \hline
		estimated    & 0.18 & 0.92 \\ \hline
	\end{tabular}
	\end{center}
	\caption{Comparison between the estimated (internal) uncertainty and the uncertainty obtained by the ground truth (external)}
	\label{tab:uncertainty}
\end{table}

\section{Conclusion}\label{sec:Conclusion}
In this paper, we proposed a Maximum Likelihood solution to the PnP problem\footnote[1]{The Matlab and C++ implementations are available at \url{https://github.com/urbste}}.
First, the variance propagation from 2D image observations to general 3D bearing vectors was introduced.
The singularity of the resulting 3$\times$3 covariance matrices of such bearing vectors motivated the subsequent reduction of the covariance to the vector tangent space.
In addition, this reduced formulation allows to obtain a solution to the PnP problem in terms of a linear Maximum Likelihood estimation followed by a fast iterative Gauss-Newton refinement.
Finally the method was tested and evaluated against all state-of-the-art PnP methods.
It shows similar execution times compared to the fastest methods and outperforms the state-of-the-art in terms of accuracy.
\section*{Acknowledgment}
This project was partially funded by the DFG research group FG 1546 "Computer-Aided Collaborative Subway Track Planning in Multi-Scale 3D City and Building Models".
In addition the authors would like to thank Patrick Erik Bradley for helpful discussions and insights.

\begin{spacing}{0.8} 
	{\small \bibliography{MLPnP_congress}}
\end{spacing}

\end{document}